\documentclass[letterpaper, 10 pt, conference]{ieeeconf}  % Comment this line out if you need a4paper

\IEEEoverridecommandlockouts                              % This command is only needed if 
\title{\LARGE \bf
Measurement-Calibrated Multi-Camera Fusion for Vision-Based Indoor Localization
}

\author{Mateo Toro Diz$^{1}$, Jonathan Hoss$^{2}$, Noah Klarmann$^{3}$% <-this % stops a space
\thanks{$^{1}$ Mateo Toro Diz is with Rosenheim Technical University of Applied Sciences,
        {\tt\small mateo.toro-diz@th-rosenheim.de}}%
\thanks{$^{2}$ Jonathan Hoss is with Rosenheim Technical University of Applied Sciences,
        {\tt\small jonathan.hoss@th-rosenheim.de}}%
\thanks{$^{3}$ Noah Klarmann is with Rosenheim Technical University of Applied Sciences,
        {\tt\small noah.klarmann@th-rosenheim.de}}%
}

\usepackage{graphicx}
\usepackage{caption}
\usepackage{acro}
\usepackage{amsmath}
\usepackage{url}

\DeclareAcronym{RMSE}{
  short = RMSE,
  long  = Root Mean Square Error
}
\DeclareAcronym{MAE}{
  short = MAE,
  long  = Mean Absolute Error
}
\DeclareAcronym{KF}{
  short = KF,
  long  = Kalman Filter
}
\DeclareAcronym{ips}{
  short = IPS,
  long  = Indoor Positioning Systems
}

\DeclareAcronym{cnn}{
  short = CNN,
  long = Convolutional Neural Network
}

\begin{document}

\maketitle
\thispagestyle{empty}
\pagestyle{empty}

%%%%%%%%%%%%%%%%%%%%%%%%%%%%%%%%%%%%%%%%%%%%%%%%%%%%%%%%%%%%%%%%%%%%%%%%%%%%%%%%
\begin{abstract}

Indoor vision-based localization systems are affected by detection noise, occlusions, and limited camera coverage, leading to uncertainty at multiple stages of the pipeline. While multi-camera data fusion is widely used to mitigate these issues, it is typically treated as a black-box component and evaluated solely end-to-end, obscuring its mechanistic contributions. To address this gap, this work investigates whether explicitly characterizing single-camera localization errors can be leveraged to calibrate and optimize multi-camera data fusion.

We introduce a measurement-calibrated fusion approach that integrates component-wise error quantification—specifically isolating homography calibration, human detection, and motion tracking. A component-wise evaluation is conducted to quantify error contributions from homography calibration, human detection, and motion tracking. 

Experimental results show that data fusion improves localization accuracy compared to single-camera baselines. While measurement-calibrated fusion provides only limited improvement in absolute accuracy over standard fusion, it substantially reduces trajectory variance and improves motion smoothness, which are critical for applications requiring stable and continuous motion estimates. These results highlight the value of explicit error characterization when designing data fusion strategies for vision-based indoor positioning systems.

\end{abstract}

%%%%%%%%%%%%%%%%%%%%%%%%%%%%%%%%%%%%%%%%%%%%%%%%%%%%%%%%%%%%%%%%%%%%%%%%%%%%%%%%
\section{INTRODUCTION}

Computer vision based localization systems are a common solution to the \ac{ips} problem, often applied to navigation and surveillance \cite{morar_comprehensive_2020}. Beyond positional accuracy, downstream applications depend on the temporal quality of trajectories, where excessive jitter, abrupt velocity changes or discontinuities can degrade navigation logic, visualization and decision-making. These issues can be partially avoided by the use of multiple cameras, which are more likely to achieve full coverage of the area of interest. Detections from multiple cameras must be fused without duplication, requiring a data fusion algorithm. 

In the literature, the implementation of computer vision based indoor localization has been studied repeatedly, both with single \cite{wu_indoor_2022,sevastopoulos_object_2020,nocera_rgbd_2025} and multiple \cite{kwon_feasibility_2023,shim_mobile_2015,carro-lagoa_alternatives_2021, cosma_camloc_2019,zaccaria_multi-robot_2021, sun_see-your-room_2019} camera systems.

Most multi-camera localization systems follow a comparable high-level pipeline, comprising object detection, projection of image coordinates into a world reference frame via homography, and a subsequent fusion stage. While implementations differ in the choice of detection networks, calibration procedures, and tracking algorithms, the overall architectural structure remains largely consistent across studies, including the present work. As a result, reported performance differences may arise from both component selection and parameterization, but are rarely analyzed at the level of individual pipeline stages.

Ground truth annotation strategies differ considerably across studies, which directly affects the comparability of reported localization errors. Shim and Cho \cite{shim_mobile_2015}, Carro-Lagoa et al. \cite{carro-lagoa_alternatives_2021} and Cosma et al. \cite{cosma_camloc_2019}, employ frame-wise two-dimensional ground truth, enabling point-to-point spatial and temporal error evaluation. Kwon et al. \cite{kwon_feasibility_2023} use manually annotated positions at discrete time intervals, introducing temporal discretization. In contrast, Zaccaria et al. \cite{zaccaria_multi-robot_2021} evaluate human localization using a one-dimensional trajectory-based ground truth, measuring only the spatial deviation from a predefined path without explicit temporal alignment. The choice of ground truth formulation determines whether the reported error reflects instantaneous 2D localization accuracy or deviation from a reference trajectory, and therefore limits direct numerical comparison across studies. In this work, we adopt the trajectory-based evaluation scheme for human localization, enabling direct comparison with Zaccaria et al. \cite{zaccaria_multi-robot_2021} while acknowledging the reduced dimensionality of the reported error metric.

The precision of the positioning systems is mostly evaluated by the comparison of the detections to a measured ground truth, presenting the error either as \ac{RMSE} or \ac{MAE}. The precision varies depending on the test setup, with values usually ranging from 10-20 cm for simple test spaces \cite{zaccaria_multi-robot_2021, sun_see-your-room_2019} and up to 1 m for larger test spaces \cite{kwon_feasibility_2023}. The best results are presented by Shim and Cho, with an error within 7.1 cm for the static case, but no information provided on the trajectory's accuracy.

However, in most existing works evaluation is performed end-to-end, where the entire system is assessed as a single unit. Kwon et al., Carro-Lagoa et al., Cosma et al., Zaccaria et al. \cite{kwon_feasibility_2023, carro-lagoa_alternatives_2021, cosma_camloc_2019, zaccaria_multi-robot_2021} do not explicitly isolate the error contribution of individual components such as homography calibration and detection, or fusion. Only Shim and Cho \cite{shim_mobile_2015} evaluate the homography separately, albeit under different conditions than the moving-object scenario. Consequently, even when benchmark datasets are used, improvements in localization accuracy cannot be attributed to specific pipeline stages, and fusion performance is rarely assessed relative to single-camera baselines. This limits the understanding of when and why fusion improves localization. 

The most decisive difference between existing systems lies in their data fusion strategies. Shim and Cho, Cosma et al., Sun et al. and Carro-Lagoa et al. \cite{shim_mobile_2015, carro-lagoa_alternatives_2021, cosma_camloc_2019, sun_see-your-room_2019} employ heuristic fusion schemes, typically combining detections through predefined equations. In contrast, Kwon et al. and Zaccaria et al. \cite{kwon_feasibility_2023, zaccaria_multi-robot_2021} utilize Kalman Filter-based fusion frameworks and provide detailed descriptions of detection handling and state estimation. However, in these works the tuning and characterization of measurement and process noise are not explicitly reported. Consequently, fusion is effectively treated as a black-box component, with limited insight into how camera-specific uncertainties influence the estimation process. Moreover, fusion accuracy is rarely evaluated relative to single-camera baselines, making it unclear whether observed improvements stem from redundancy, temporal smoothing, bias compensation, or mere reduction of aggregate error magnitude. 

This work investigates if the measurement of single camera error can be used to calibrate data fusion by evaluating the system's component individually, and the effect of measurement-calibrated fusion on the localization precision and trajectory smoothness.

We present the following contributions:
\begin{itemize}

\item Component-wise evaluation of a vision-based localization pipeline.
\item We propose a covariance-calibrated measurement noise assignment strategy for Kalman Filter-based fusion
\item We demonstrate that explicitly accounting for camera-specific error characteristics reduces trajectory variance and improves motion smoothness, thereby better satisfying requirements for real-time continuous tracking applications.

\end{itemize}

\section{Methodology}

\subsection{System overview}
This paper presents a multi-camera object detection and localization system that combines neural networks to detect and classify objects, projective transformation for spatial localization, and data fusion to improve localization accuracy and temporal consistency. The complete source code of the proposed system is available at \url{github.com/proto-lab-ro/measurement-calibrated-data-fusion}.

Spatial localization is performed using homography, which maps image pixels to ground-plane positions. Objects are first detected with a \ac{cnn}, and the resulting detections are fed to a pose estimation network. When pose estimation is successful, the position of the person is computed from the detected foot keypoints. If pose estimation fails, the bottom-center of the \ac{cnn} bounding box is used as a fallback estimate.

YOLO v8n \cite{yaseen_what_2024} is employed for object detection. MediaPipe \cite{bazarevsky_blazepose_2020} is used for 2D human pose estimation. The pose estimation network predicts a set of anatomical keypoints (e.g., ankles, knees, hips) in image coordinates for each detected person. In this work, the ankle keypoints are used to estimate the subject’s ground contact position, improving spatial localization compared to using the bounding box alone.

A linear \ac{KF} fuses detections coming from the individual camera systems, providing temporal continuity and explicit control over trajectory smoothness through process and measurement noise modeling. To aid the data fusion algorithm, all three cameras are synchronized using Network Time Protocol (NTP). NTP aligns the internal clocks of distributed devices over a network, ensuring detections from different cameras correspond to the same physical instant.

Standard and measurement-calibrated fusion are compared. In standard data fusion, only the process and measurement models are provided to handle incoming detections. A more detailed explanation of the corresponding algorithm is given in the next subsection.

Measurement-calibrated data fusion accounts for  the consistency of the camera’s detections when the test subject is stationary to adjust the confidence assigned by the \ac{KF} to the incoming measurements. In addition, it incorporates knowledge of each camera’s effective field of view, allowing detections caused by artifacts or interpolation to be discarded.

\subsection{Data fusion algorithm}
The data fusion algorithm is based on a linear \ac{KF}, which combines the detections from multiple cameras into a single, consistent trajectory to ensure temporal continuity across coverage gaps.

The filter state at time step $k$ is defined as follows:
$$
\mathbf{x}_k =
\left[
\begin{array}{cccc}
x_k & v_{x,k} & y_k & v_{y,k}
\end{array}
\right]
\eqno{(1)}
$$
where $x_k$ and $y_k$ denote the target position in world coordinates, and $v_{x,k}$ and $v_{y,k}$ the corresponding velocity components.

State prediction follows a discrete-time constant-velocity motion model:
$$
\mathbf{x}_{k|k-1}
=
\mathbf{F}\,\mathbf{x}_{k-1}
\eqno{(2)}
$$
with $\mathbf{F}$, of size \(4 \times 4\), the state transition matrix. Deviations from linear motion are modeled as process noise, represented by the process noise matrix $\mathbf{Q}$ in the predicted state covariance:
$$
\mathbf{P}_{k|k-1}
=
\mathbf{F}\,\mathbf{P}_{k-1}\,\mathbf{F}^{\mathsf{T}} + \mathbf{Q}
\eqno{(3)}
$$
The process noise matrix $\mathbf{Q}$, of size \(4 \times 4\), is tuned via random search by minimizing the \ac{MAE} on validation data.

Each camera provides 2D keypoint positions in the world coordinates. The measurement model is linear:
$$
\mathbf{z}_k(i) = H \mathbf{x}_k + \mathbf{v}_k(i), \quad \mathbf{v}_k(i) \sim \mathcal{N}(0, \mathbf{R}(i))
\eqno{(4)}
$$
where $\mathbf{H}$, of size \(2 \times 4\) maps the state vector to position measurements, as only positions are directly observed:
$$
\mathbf{H} =
\left[
\begin{array}{cccc}
1 & 0 & 0 & 0 \\
0 & 0 & 1 & 0
\end{array}
\right]
$$

The matrix $\mathbf{R}(i)$, of size \(2 \times 2\) is the measurement noise covariance matrix for detection $i$.
For standard fusion the matrix $\mathbf{R}(i)$ is constant, derived from a fixed measurement variance $ \mathbf \sigma^2_{standard} $ applied to all cameras and positions.

We avoid using a single confidence score because camera measurement errors differ between the x- and y-axes. Furthermore, relying solely on field-of-view gating is insufficient, as it only improves tracking at the edges of the camera's vision.

In measurement-calibrated fusion, $\mathbf{R}(i)$ is assigned based on results from static-camera tests (Section \ref{section:hom_static_tests}). For each detection, we identify the nearest static test point and assign the corresponding measurement noise. This adaptive assignment modifies the \ac{KF} gain for each detection: high-precision measurements are weighted more strongly, while low-precision ones are down-weighted. Although calibration is based on static measurements, we assume that spatially varying measurement variance primarily reflects camera geometry, detection uncertainty and projection error, which remain consistent during motion. Motion related deviations are modeled through the process noise term of the \ac{KF}.

Data association is handled via Joint Probabilistic Data Association \cite{bar-shalom_probabilistic_2009}, implemented using the $StoneSoup$ Python Library \cite{hiscocks_stone_2023}. The algorithm is not modified; only pre-gating is introduced, discarding detections outside the validated field of view of each camera to prevent artifacts from influencing the state update.

Finally, to compensate for detection latency across cameras, detections are time-binned in \(0.08\,\text{s}\) windows to treat them as simultaneous observations.

Overall, measurement-calibrated \ac{KF} fusion accounts for camera-specific error characteristics to modify the confidence of the individual detections, following accurate detections more closely and avoiding getting distracted by low-accuracy detections.

\subsection{Test environment and coordinate system}
\begin{figure*}[t]
\centering
\includegraphics[width=\columnwidth]{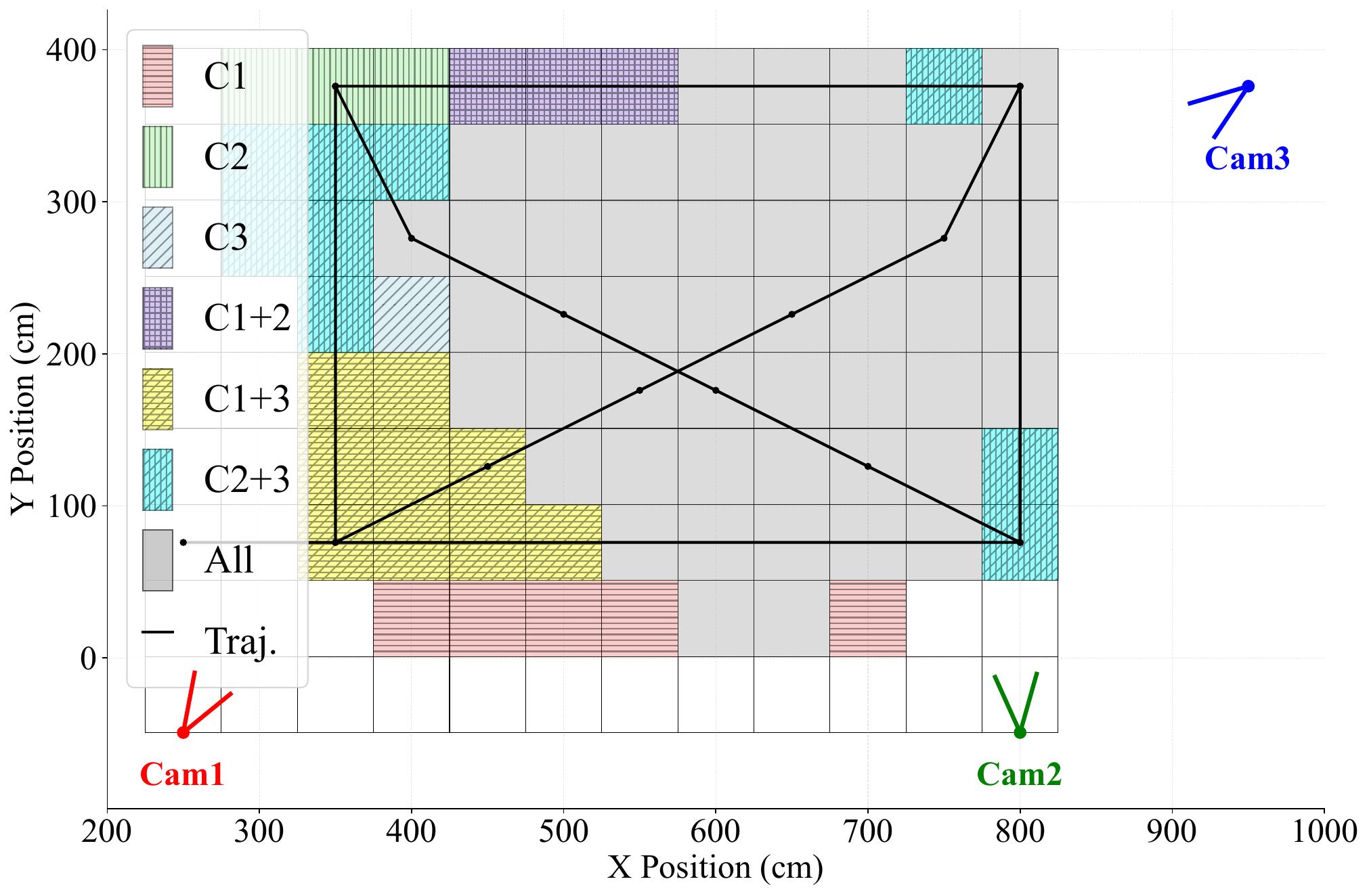}
\caption{Top-down view of the reconstructed trajectory for a single-camera setup.}
\label{fig:coverage_map}
\end{figure*}

The data used to evaluate the improvements brought by the fusion of data from multiple static cameras was collected by means of a dynamic movement experiment, in a $550x300  cm$ section of the test floor space. The movement of a single person was captured by three units of the single camera system, placed at the edges of the measurement area, as shown in Fig. \ref{fig:coverage_map}. The cameras are mounted such that the union of their coverage areas spans the entire shop floor, while none of the single cameras covers the full test space. This setup enables  evaluation of the effect of the number of cameras on the resulting system precision.

The coordinate system is Cartesian, with all movement constrained to the horizontal plane. This plane is also used to calibrate the homography, which maps detections from camera image coordinates to the real-world coordinate system. Calibration is performed using a total of 24 points. The effect of using a grid for homography calibration was evaluated and discarded, as it did not lead to a significant improvement in localization accuracy.

\subsection{Single camera error characterization} \label{section:hom_static_tests}
Uncertainty is gradually introduced into the pipeline by systematically isolating the error contributions of individual components across two testing stages. First, the homography error is isolated by assuming the pose estimation algorithm provides the exact pixel location of the test subject, such that any observed localization error can be attributed solely to the homography.

The second stage of  systematic testing introduces an additional layer of complexity by measuring the error of the system when estimating the static position of a person. This allows for the characterization of the single camera system error as a function of the test subject's position. The resulting error standard deviation  is used to parameterize the measurement noise of the measurement model used by the \ac{KF} for data fusion.

\subsection{Dynamic evaluation and metrics}
The ground truth is defined as a set of straight-line segments connecting known waypoints, which the test subject follows during the experiment. The trajectory is designed to cover a representative portion of the floor surface while maintaining a clear geometric shape. This layout also enables the evaluation of the effect of camera coverage on the system's performance.

The improvements introduced by data-fusion are measured in terms of both raw localization precision and motion smoothness. Localization accuracy is quantified using  \ac{MAE} and \ac{RMSE}. Motion smoothness is evaluated as it is a critical requirement for real-time tracking and visualization of objects.

The dynamic test is conducted by moving along the defined trajectory with approximately constant speed. To reduce the influence of human error, the experiment was repeated eleven times, from which nine runs without disruptions were retained for analysis. The system exhibits consistent behavior across these runs, with the standard deviation of \ac{RMSE} within two centimeters.

Global system accuracy is measured by comparing the estimated trajectory with the ground truth, represented as a set of segments. This comparison is performed using a  point-to-segment distance calculation, which yields the cross-track error. The error is evaluated with \ac{MAE} and \ac{RMSE}, allowing comparison with other related works in the literature \cite{carro-lagoa_alternatives_2021,cosma_camloc_2019,zaccaria_multi-robot_2021}.

Axis-dependent accuracy is evaluated only along the outermost segments of the trajectory, as diagonal motion is influenced by the accuracy of both $x$ and $y$ axis, and therefore does not provide a clear separation of axis-specific accuracy.

Similarly, the effect of camera coverage on localization accuracy is assessed exclusively in the outermost portion of the trajectory, as these regions are not simultaneously covered by all three cameras, unlike the diagonal segments. 

The smoothness of the estimated trajectory is quantified with a custom metric defined as the standard deviation of the inter-sample displacement. Since all cameras and fusion methods operate at the same sample rate, this metric directly reflects short-term motion variability. Higher displacement variance indicates jitter and abrupt corrections, while lower values correspond to smoother, more temporally consistent trajectories.

\section{Results}

\subsection{Quantitative effect of data fusion}
The effect of data fusion is evaluated by comparing the point-to-segment error for single-camera localization, standard data fusion, and measurement-calibrated data fusion, as reported in Table \ref{table_metrics}. Both fusion approaches reduce localization error relative to the best-performing single camera, achieving a reduction of at least 6\% in \ac{RMSE} and a similar decrease in \ac{MAE}.

Compared to standard fusion, measurement-calibrated data fusion yields a further reduction in \ac{RMSE} of 6\%, while the \ac{MAE} remains comparable between the two fusion methods.

The most pronounced benefit of measurement-calibrated fusion is observed in trajectory smoothness. The displacement standard deviation is reduced by approximately 50\% relative to two of the individual cameras, indicating a substantially smoother estimated trajectory.

\begin{table}[b]
\caption{Localization error metrics for single-camera and fusion methods.}
\label{table_metrics}
\begin{center}
\begin{tabular}{|c||c|c|c|c|c|}
\hline
metric & cam1& cam2& cam3& standard &calibrated \\
\hline
RMSE (cm)     & 15.78 & 17.62 & 19.04 & 14.78 & 13.88 \\
\hline
MAE (cm)      & 12.74 & 13.40 & 15.20 & 11.58 & 11.10 \\
\hline
disp\_std (cm) & 15.19 & 13.19 & 15.10 & 8.53  & 7.71  \\
\hline
\end{tabular}
\end{center}
\end{table}

\subsection{Single camera error characterization results}
The measurement-calibrated data fusion algorithm is supported by stepwise evaluation of the localization pipeline. Table \ref{table_error_components} reports the per-camera error contributions of homography, human detection, and movement tracking.

The homography error remains bounded, with values below six~cm for all cameras and axes. Incorporating human detection increases the total localization error and introduces axis-dependent asymmetries for most cameras, with noticeable differences between the $x$ and $y$ directions. These asymmetries are reduced and, in some cases, reversed during dynamic tests, indicating the presence of additional motion-related error sources.

\begin{table}[b]
\caption{Per-camera localization error components for different estimation methods.}
\label{table_error_components}
\begin{center}
\begin{tabular}{|c||c|c|c|}
\hline
 & cam1 & cam2 & cam3 \\
\hline
homography-x (cm) & 3.25 & 2.49 & 6.10 \\
\hline
homography-y (cm) & 2.73 & 3.25 & 3.96 \\
\hline
static-x (cm) & 13.19 & 9.31 & 7.38 \\
\hline
static-y (cm) & 3.63 & 11.70 & 5.55 \\
\hline
dynamic-x (cm) & 11.46 & 10.07 & 19.71 \\
\hline
dynamic-y (cm) & 14.36 & 17.18 & 13.22 \\
\hline
\end{tabular}
\end{center}
\end{table}

\subsection{Qualitative effects of data fusion}
\begin{figure*}[t]
\centering
\includegraphics[width=\columnwidth]{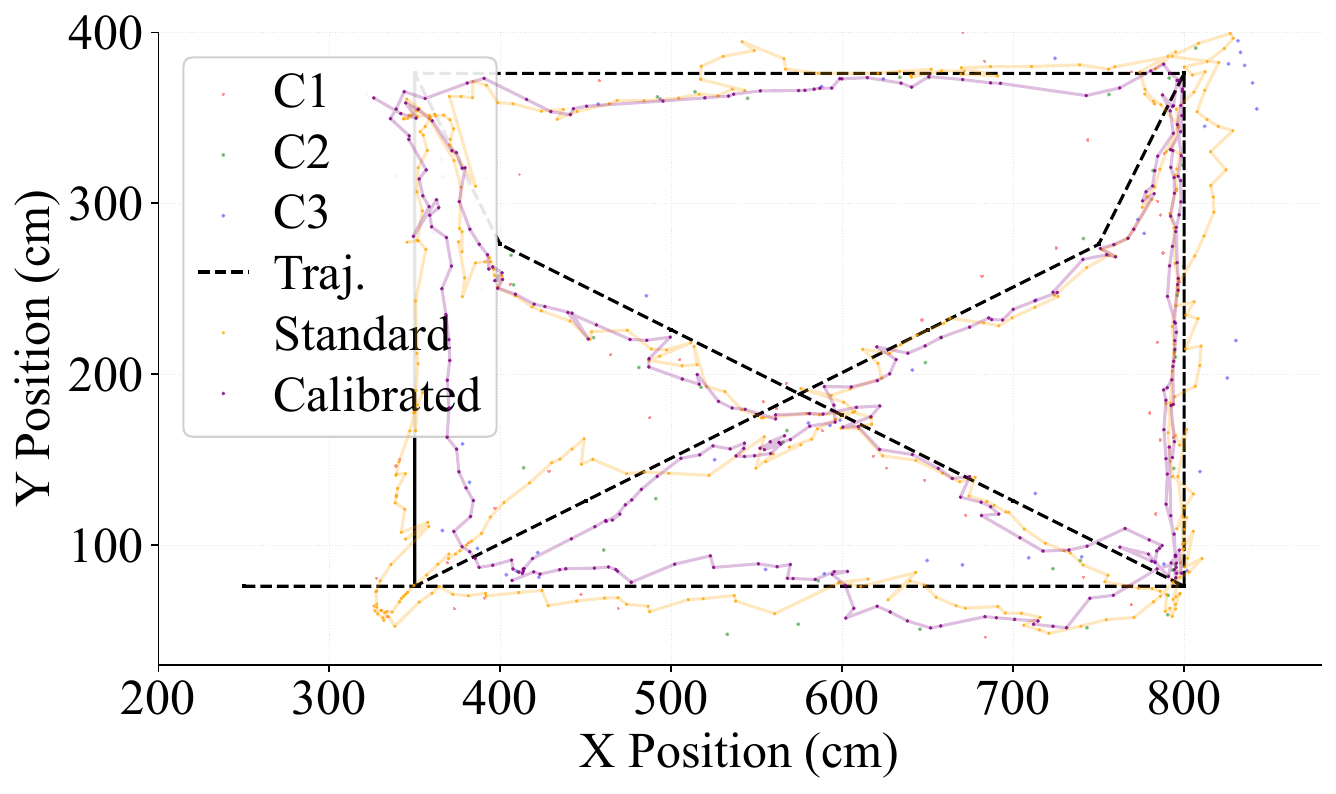}
\caption{Overlay of the ground truth, raw detections, standard and calibrated fusion tracks.}
\label{fig:trajectory_single}
\end{figure*}

A secondary but relevant benefit of data fusion is the ability to maintain trajectory continuity across single-camera coverage gaps. As none of the three cameras provides full coverage of the shop floor, single-camera tracking is intermittently lost and affected by noisy detections. The use of data fusion, and in particular measurement-calibrated fusion, results in a visually smoother trajectory, as illustrated in Fig. \ref{fig:trajectory_single}, where the ground truth, raw detections, and calibrated fusion tracks are overlaid. The fusion track mitigates the variability of the raw detections, yielding a smoother approximation of the object’s true motion.

\subsection{Effect of camera coverage in the system's accuracy}
\begin{figure}[t]
\centering
\includegraphics[width=\columnwidth]{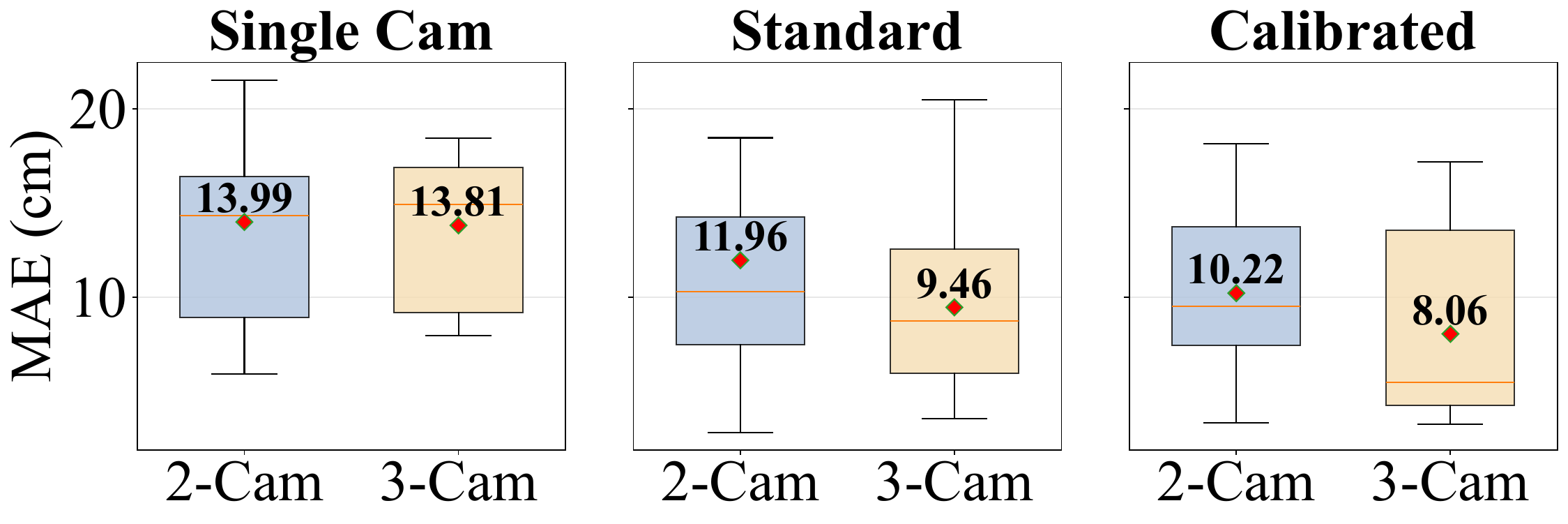}
\caption{Comparison of the effect of number of cameras covering test area for single camera and fusion systems. }
\label{fig:coverage_boxplot}
\end{figure}

Lastly, the effect of camera coverage on localization accuracy is evaluated for both data fusion and single-camera systems. The \ac{RMSE} is compared between trajectory segments where all three cameras provide coverage and segments where at least one camera does not, with each segment treated as an individual data point. The resulting error distributions are shown in Fig. \ref{fig:coverage_boxplot}.

For single-camera systems, variations in camera coverage do not significantly affect the error magnitude, although the spread of the error distribution increases as coverage decreases. In contrast, data fusion benefits from increased camera coverage, with the \ac{RMSE} decreasing by at least two~cm for both the standard and calibrated fusion methods when all cameras contribute.

This reduction in error is associated with the increased number of observations available to the fusion algorithm, which mitigates the camera-specific error characteristics reported in Table \ref{table_error_components}. These results indicate that, within the evaluated setup, the accuracy gains from data fusion increase with the number of available camera observations.

\section{Discussion and limitations}
This work investigates under which conditions multi-camera data fusion improves vision-based indoor localization, beyond end-to-end accuracy comparisons. By evaluating the localization pipeline component-wise, we characterize how the system’s behavior evolves as individual components are introduced and how their error characteristics influence fusion performance.

The results show that standard data fusion consistently improves localization accuracy relative to single-camera baselines, primarily through redundancy and temporal smoothing. In contrast, informing the fusion process with empirically measured camera uncertainty yields only modest improvements in absolute accuracy, suggesting that overall performance is largely bounded by upstream components such as object detection and homography calibration.

Despite limited improvements in absolute error, measurement-calibrated fusion produces a pronounced reduction in trajectory variance and substantially smoother tracks. This behavior arises from the adaptive tuning of the measurement noise covariance, which influences the Kalman gain through the probabilistic estimation process. As a result, measurement-calibrated fusion yields more stable and robust trajectories, which is particularly relevant for real-time systems where downstream modules implicitly assume smooth motion and may amplify noise-induced fluctuations.

These findings address a gap in the literature, where data fusion is often treated as a black-box component and evaluated primarily through end-to-end metrics. By explicitly linking camera-specific error characteristics to fusion behavior, this work provides a principled basis for the design and evaluation of multi-camera localization systems.

In terms of numerical accuracy, the reported results are comparable to those in the literature, achieving a significantly lower error than Kwon et al., Carro-Lagoa et al., and Cosma et al. \cite{kwon_feasibility_2023, carro-lagoa_alternatives_2021, cosma_camloc_2019}. Zaccaria et al. \cite{zaccaria_multi-robot_2021} achieves slightly lower error results in its second experiment, where the error is also measured as the distance between a segment and the person. However, the trajectory is a single line, whereas in this paper a more complex trajectory is followed. Lastly, Sun et al. \cite{sun_see-your-room_2019} achieves a higher accuracy when the test subject is between 0-3 meters from the camera, but similar results once the test subject is further away.

While several works employ multi-camera data fusion \cite{kwon_feasibility_2023, cosma_camloc_2019, zaccaria_multi-robot_2021}, none explicitly evaluates the effect of informing fusion with empirically measured camera uncertainty.

Several limitations must be acknowledged. First, ground truth is available only as a single-dimensional segment, which restricts direct comparison with two-dimensional ground truth papers. Second, the evaluation consists of a single tracked subject, which matches most of the literature \cite{shim_mobile_2015, carro-lagoa_alternatives_2021, cosma_camloc_2019, sun_see-your-room_2019} but does not consider multi-target conditions. Finally, the experiments are conducted along a straight-line trajectory, which favors the linear assumption of the \ac{KF}.

Future work should extend this evaluation framework to settings with higher-fidelity ground truth, such as motion-capture systems, enabling full two-dimensional error analysis. Additionally, evaluating multi-target tracking scenarios would further align the approach with practical deployment conditions.

\addtolength{\textheight}{-12cm}  

%%%%%%%%%%%%%%%%%%%%%%%%%%%%%%%%%%%%%%%%%%%%%%%%%%%%%%%%%%%%%%%%%%%%%%%%%%%%%%%%

%%%%%%%%%%%%%%%%%%%%%%%%%%%%%%%%%%%%%%%%%%%%%%%%%%%%%%%%%%%%%%%%%%%%%%%%%%%%%%%%

%%%%%%%%%%%%%%%%%%%%%%%%%%%%%%%%%%%%%%%%%%%%%%%%%%%%%%%%%%%%%%%%%%%%%%%%%%%%%%%%

\bibliographystyle{ieeetr}
\bibliography{MOCAS}

\end{document}